# A Neuro-Fuzzy Approach for Modelling Electricity Demand in Victoria

## Ajith Abraham and Baikunth Nath

School of Computing and Information Technology
Monash University (Gippsland Campus),
Churchill, Australia 3842
Email: {Ajith.Abraham, Baikunth.Nath}@infotech.monash.edu.au

## Abstract

Neuro-Fuzzy systems have attracted growing interest of researchers in various scientific and engineering areas due to the increasing need of intelligent systems. This paper evaluates the use of two popular soft computing techniques and conventional statistical approach based on Box-Jenkins Autoregressive Integrated Moving Average (ARIMA) model to predict electricity demand in the State of Victoria, Australia. The soft computing methods considered are an evolving fuzzy neural network and an artificial neural network trained using scaled conjugate gradient algorithm and backpropagation algorithm. The forecast accuracy is compared with the forecasts used by Victorian Power Exchange (VPX) and the actual energy demand. To evaluate, we considered load demand patterns for ten consecutive months taken every 30 minutes for training the different prediction models. Test results show that the neuro fuzzy system performed better than neural networks, ARIMA model and the VPX forecasts.

## 1. Introduction and Related Research

Traditionally, the energy sector, and particularly the electricity sector, has been dominated by monopoly or near monopoly enterprises, typically either owned or regulated by government. The nature of electricity market is changing very rapidly with a widespread international movement towards competitiveness. Some countries, such as Norway, Chile, Japan, UK and the United States have commonly been supplied electricity by a large number of different regional Generators and have developed a variety of mechanisms to allow some form of trade between them. In 1994 Victoria started the process of privatisation and restructuring electricity industry to generate competition. The objective was to promote a more flexible, cost-effective and efficient electricity industry with the aim of delivering cheaper electricity to business and the general community. Following success of this operation, Australia started the process of implementing a unified National Electricity Market in December 1998 [13].

In Victoria, more than 80 percent of electricity comes from brown coal fired stations. Since the on/off states for such power generating stations have large time lags for start-ups, stability in its operations is important. The peak demand for electricity for the State at any time instant is about 6700 MWh. This demand is highly volatile on a day-to-day basis and is being significantly affected by the Victorian weather conditions. Electricity is consumed as it is generated and is very often sold in advance of production. Electricity as a commodity has very different characteristics compared to a physical commodity. It cannot be stored. Therefore, to meet the electricity market demands a highly reliable supply and delivery system is required. Additionally, in order to gain a competitive advantage in this market through the spot-market pricing an accurate forecast of electricity demand at regular time intervals is essential. Until 1996, Victorian Power Exchange (VPX) the body responsible for the secure operations of the power system, generated electricity demand forecasts based on weather forecasts and historical demand patterns [1] [2]. Our research is focused on developing more accurate and reliable forecasting models that improves current forecasting methods. This paper investigates two soft computing forecasting models and a popular statistical forecasting technique based on Box - Jenkins ARIMA model [12] for predicting 96 half-hourly (two days ahead) demands for electricity, and compares their performance with forecasts used by VPX. The soft computing models considered are an

Evolving Fuzzy Neural Network (EFuNN) [8] and a feedforward Artificial Neural Network (ANN) trained using the scaled gradient conjugate algorithm and backpropagation algorithm.

For developing the forecasting models we used the energy demand data for ten months period from 27[th] January to 30[th] November 1995 in the State of Victoria. We also made use of the associated data stating the minimum and maximum temperature of the day, time of day, season and the day of week. The forecasting models were trained using 3 randomly selected samples containing 20% of the data during the period 27[th] January 1995 to 28 November 1995. To ascertain the forecasting accuracy the developed models were tested to predict the demand for the period (29-30) November 1995.

The paper is organised as follows. In Section 2, we give a brief overview of EFuNN and ANNs and ARIMA model. Section 3 discusses the experimentation set-up, characteristics of the data and the forecasting performance of the proposed neuro fuzzy system, artificial neural networks and ARIMA model. Conclusions are drawn in Section 4.

## 2. Forecasting Models

A wide variety of forecasting methods are available to the management. The evolution of soft computing techniques has increased the understanding of various aspects of the problem environment and consequently the predictability of many events. Connectionist models [11] make use of some of the popular soft computing techniques including methods of neurocomputing [7], neuro-fuzzy computing [3], evolutionary algorithms and several hybrid techniques. In contrast with the conventional AI techniques, which deal only with precision, certainty and rigor, connectionist models are able to exploit the tolerance for imprecision, uncertainty and are often very robust.

Perhaps no other statistical forecasting technique has been more widely discussed than ARIMA model building. An ARIMA model has three components: AutoRegressive, Integrated and Moving Average. Basic ARIMA model building consists of four steps: (1) model identification (2) parameter estimation (3) model diagnostics (4) forecast verification and reasonableness. We hope the following sections will give some technical insights of the different prediction models considered.

### 2.1 Artificial Neural Network (ANN)

Artificial neural networks were designed to mimic the characteristics of the biological neurons in the human brain and nervous system [7]. An artificial neural network creates a model of neurons and the connections between them, and trains it to associate output neurons with input neurons. The network "learns" by adjusting the interconnections (called weights) between layers. When the network is adequately trained, it is able to generate relevant output for a set of input data. A valuable property of neural networks is that of generalisation, whereby a trained neural network is able to provide a correct matching in the form of output data for a set of previously unseen input data.

Backpropagation (BP) is one of the most famous training algorithms for multilayer perceptrons [8]. Basically, BP is a gradient descent technique to minimize the error $E$ for a particular training pattern. For adjusting the weight ($w_k$), in the batched mode variant the descent is based on the gradient $\nabla E$ ($\frac{\delta E}{\delta w_k}$) for the total training set:

$$\Delta w_k(n) = -\varepsilon \cdot \frac{\delta E}{\delta w_k} + \alpha \cdot \Delta w_k(n-1) \qquad (1)$$

The gradient gives the direction of error $E$. The parameters $\varepsilon$ and $\alpha$ are the learning rate and momentum respectively. A good choice of both the parameters is required for training success and speed of the ANN.

In the Conjugate Gradient Algorithm (CGA) a search is performed along conjugate directions, which produces generally faster convergence than steepest descent directions [10]. A search is made along the conjugate gradient direction to determine the step size, which will minimize the performance function along that line. A line search is performed to determine the optimal distance to move along the current search direction. Then the next search direction is determined so that it is conjugate to previous search direction. The general procedure for determining the new search direction is to combine the new steepest descent direction with the previous search direction. An important feature of the CGA is that the minimization performed in one step is not partially undone by the next, as it is the case with gradient descent methods. An important drawback of CGA is the requirement of a line search, which is computationally expensive. Moller [9] introduced the Scaled Conjugate Gradient Algorithm (SCGA) as a way of avoiding the complicated line search procedure of conventional CGA. According to the SCGA, the Hessian matrix is approximated by

$$E''(w_k)p_k = \frac{E'(w_k + \sigma_k p_k) - E'(w_k)}{\sigma_k} + \lambda_k p_k \qquad (2)$$

where $E'$ and $E''$ are the first and second derivative information of global error function $E(w_k)$. The other terms $p_k$, $\sigma_k$ and $\lambda_k$ represent the weights, search direction, parameter controlling the change in weight for second derivative approximation and parameter for regulating the indefiniteness of the Hessian. In order to get a good quadratic approximation of $E$, a mechanism to raise and lower $\lambda_k$ is needed when the Hessian is positive definite. Detailed step-by-step description can be found in [9].

## 2.2 Neuro-Fuzzy Network

We define a neuro-fuzzy system [3] as a combination of ANN and Fuzzy Inference System (FIS) [6] in such a way that neural network learning algorithms are used to determine the parameters of FIS [4]. An even more important aspect is that the system should always be interpretable in terms of fuzzy *if-then* rules, because it is based on the fuzzy system reflecting vague knowledge. We used Evolving Fuzzy Neural Network (EFuNN) implementing a Mamdani [5] type FIS and all nodes are created during learning. EFuNN has a five-layer structure as shown in Figure 1 [8].

Figure 2 illustrates a Mamdani FIS combining 2 fuzzy rules using the *max-min* method [5]. According to the Mamdani FIS, the rule antecedents and consequents are defined by fuzzy sets and has the following structure:

*if $x$ is $A_1$ and $y$ is $B_1$ then $z_1 = C_1$* $\qquad (3)$

where $A_1$ and $B_1$ are the fuzzy sets representing input variables and $C_1$ and is the fuzzy set representing the output fuzzy set. In EFuNN, the input layer is followed by the second layer of nodes representing fuzzy quantification of each input variable space. Each input variable is represented here by a group of spatially arranged neurons to represent a fuzzy quantization of this variable. Different membership functions (MF) can be attached to these neurons (triangular, Gaussian, etc.). The nodes representing membership functions can be modified during learning. New neurons can evolve in this layer if, for a given input vector, the corresponding variable value does not belong to any of the existing MF to a degree greater than a membership threshold. The third layer contains rule nodes that evolve through hybrid supervised/unsupervised learning. The rule nodes represent prototypes of input-output data associations, graphically represented as an association of hyper-spheres from the fuzzy input and fuzzy output spaces. Each rule node, e.g. $r_j$, represents an association between a hyper-sphere from the fuzzy input space and a hyper-sphere from the fuzzy output space; $W_1(r_j)$ connection weights representing the co-ordinates of the center of the sphere in the fuzzy input space, and $W_2(r_j)$ – the co-ordinates in the fuzzy output space. The radius of an input hyper-sphere of a rule node is defined as (1- *Sthr*), where *Sthr* is the sensitivity threshold parameter defining the minimum activation of a rule node (e.g., $r_1$, previously evolved to represent a data point $(X_{d1}, Y_{d1})$) to an input vector (e.g., $(X_{d2}, Y_{d2})$) in order for the new input vector to be associated with this rule node. Two pairs of fuzzy input-output data vectors $d_1=(X_{d1}, Y_{d1})$ and $d_2=(X_{d2}, Y_{d2})$ will be allocated to the first rule node $r_1$ if they fall into the $r_1$ input sphere and in the $r_1$ output sphere, i.e. the local normalised fuzzy difference between $X_{d1}$ and $X_{d2}$ is smaller than the radius $r$ and the local normalised fuzzy difference between $Y_{d1}$ and $Y_{d2}$ is smaller than an error threshold *Errthr*. The local normalised fuzzy difference between two fuzzy membership vectors $d_{1f}$ and $d_{2f}$ that represent the membership degrees to which two real values $d_1$ and $d_2$ data belong to the pre-defined MF, are calculated as $D(d_{1f}, d_{2f}) = sum(abs(d_{1f} - d_{2f}))/sum(d_{1f} + d_{2f})$.

If data example $d_1 = (X_{d1}, Y_{d1})$, where $X_{d1}$ and $X_{d2}$ are correspondingly the input and the output fuzzy membership degree vectors, and the data example is associated with a rule node $r_1$ with a centre $r_1^1$, then a new data point $d_2=(X_{d2}, Y_{d2})$, will also be associated with this rule node through the process of associating (learning) new data points to a rule node. The centres of this node hyper-spheres adjust in the fuzzy input space depending on a learning rate $lr_1$, and in the fuzzy output space depending on a learning rate $lr_2$, on the two data point's $d_1$ and $d_2$. The adjustment of the centre $r_1^1$ to its new position $r_1^2$ can be represented mathematically by the change in the connection weights of the rule node $r_1$ from $W_1(r_1^1)$ and $W_2(r_1^1)$ to $W_1(r_1^2)$ and $W_2(r_1^2)$ according to the following vector operations:

$$W_2(r_1^2) = W_2(r_1^1) + lr_2 \cdot Err(Y_{d1}, Y_{d2}) \cdot A_1(r_1^1) \qquad (4)$$

$$W_1(r_1^2) = W_1(r_1^1) + lr_1 \cdot Ds(X_{d1}, X_{d2}) \qquad (5)$$

where $Err(Y_{d1}, Y_{d2}) = Ds(Y_{d1}, Y_{d2}) = Y_{d1} - Y_{d2}$ is the signed value rather than the absolute value of the fuzzy difference vector; $A_1(r_1^1)$ is the activation of the rule node $r_1^1$ for the input vector $X_{d2}$.

While the connection weights from $W_1$ and $W_2$ capture spatial characteristics of the learned data (centres of hyper-spheres), the temporal layer of connection weights $W_3$ captures temporal dependencies between consecutive data examples. If the winning rule node at the moment $(t-1)$ (to which the input data vector at the moment $(t-1)$ was associated) was $r_1=inda_1(t-1)$, and the winning node at the moment $t$ is $r_2=inda_1(t)$, then a link between the two nodes is established as follows:

$$W_3(r_1, r_2)^{(t)} = W_3(r_1, r_2)^{(t-1)} + lr_3 \cdot A_1(r_1)^{(t-1)} A_1(r_2))^{(t)}, \qquad (6)$$

where: $A_1(r)^{(t)}$ denotes the activation of a rule node $r$ at a time moment $(t)$; $lr_3$ defines the degree to which the EFuNN associates links between rules (clusters, prototypes) that include consecutive data examples (if $lr_3=0$, no temporal associations are learned in an EFuNN structure).

The learned temporal associations can be used to support the activation of rule nodes based on temporal, pattern similarity. Here, temporal dependencies are learned through establishing structural links. The ratio spatial-similarity/temporal-correlation can be balanced for different applications through two parameters $S_s$ and $T_c$ such that the activation of a rule node $r$ for a new data example $d_{new}$ is defined as the following vector operations:

$$A_1(r) = f(S_s \cdot D(r, d_{new}) + T_c \cdot W_3(r^{(t-1)}, r)) \qquad (7)$$

where $f$ is the activation function of the rule node $r$, $D(r, d_{new})$ is the normalised fuzzy distance value and $r^{(t-1)}$ is the winning neuron at the previous time moment.

The fourth layer of neurons represents fuzzy quantification for the output variables. The fifth layer represents the real values for the output variables.

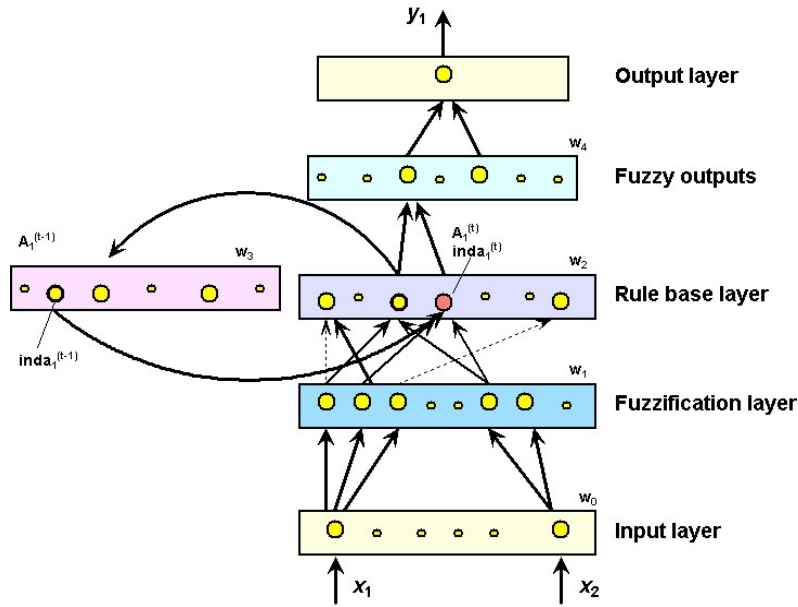

**Figure 1.** Architecture of EFuNN

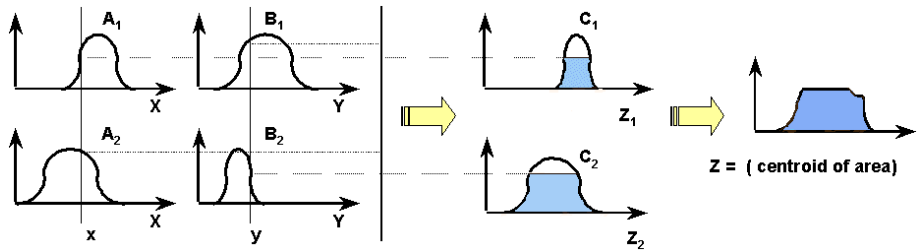

**Figure 2.** Mamdani fuzzy inference system

EFuNN evolving algorithm is given as a procedure of consecutive steps [8]:

1) Initialize an EFuNN structure with a maximum number of neurons and zero value connections. If initially there are no rule nodes connected to the fuzzy input and fuzzy output neurons, then create the first node $r_j=1$ to represent the first data example $EX= (X_{d1}, Y_{d1})$ and set its input $W_1$ ($r_j$) and output $W_2$ ($r_j$) connection weights as follows:

   <*Create a new rule node $r_j$*> to represent a data sample $EX: W_1(r_j)=EX: W_2(r_j)= TE$, where *TE* is the fuzzy output vector for the (fuzzy) example *EX*.

2) While <there are data examples> Do

   Enter the current, example $(X_{di}, Y_{di})$, *EX* being the fuzzy input vector (the vector of the degrees to which the input values belong to the input membership functions). If there are new variables that appear in this example and have not been used in previous examples, create new input and/or output nodes with their corresponding membership functions.

3) Find the normalized fuzzy similarity between the new example *EX* (fuzzy input vector) and the already stored patterns in the case nodes $r_j= r_1, r_2, ...., r_n$

$$D(EX, r_j) = \text{sum (abs } (EX - W_1(r_j))) / \text{sum } (W_1(r_j) + EX)$$

4) Find the activation $A_1(r_j)$ of the rule nodes $r_j = r_1, r_2, ...., r_n$. Here radial basis activation (*radbas*) function, or a saturated linear (*satlin*) one, can be used, i.e.

   $A_1(r_j) = \text{radbas } (S_s D(EX, r_j - T_c W_3))$, or $A_1(r_j) = \text{satlin } (1 - S_s D(EX, r_j + T_c W_3))$.

5) Update the pruning parameter values for the rule nodes, e.g. age, average activation as pre-defined.

6) Find *m* case nodes $r_j$ with an activation value $A_1(r_j)$ above a predefined sensitivity threshold *Sthr*.

7) From the *m* case nodes, find one rule node *inda₁* that has the maximum activation value *maxa₁*.

8) If *maxa₁* < *Sthr*, then, <create a new rule node> using the procedure from step 1.

   Else

9) Propagate the activation of the chosen set of *m* rule nodes $(r_{j1},...,r_{jm})$ to the fuzzy output neurons:
   $A_2 = \text{satlin } (A_1(r_{j1},...,r_{jm}) . W_2)$

10) Calculate the fuzzy output error vector: $Err = A_2 - TE$

11) If $(D(A_2, TE) > Errthr)$ <create a new rule node> using the procedure from step 1.

12) Update (a) the input, and (b) the output of the *m-1* rule nodes $k = 2 : j_m$ in case of a new node was created, or *m* rule nodes $k = j_1 : j_m$, in case of no new rule was created:

    - $Ds(EX - W_1(r_k)) = EX - W_1(r_k)$; $W_1(r_k) = W_1(r_k) + lr_1 . Ds(EX - W_1(r_k))$, where $lr_1$ is the learning rate for the first layer;

    - $A_2(r_k) = \text{satlin } (W_2(r_k).A_1(r_k))$; $Err(rk) = TE - A_2(r_k)$;

    - $W_2(r_k) = W_2(r_k) + lr_2 . Err(r_k) . A_1(r_k)$, where $lr_2$ is the learning rate for the second layer.

13) Prune rule nodes $r_j$ and their connections that satisfy the following fuzzy pruning rule to a pre-defined level representing the current need of pruning:

    IF (a rule node $r_j$ is OLD) and (average activation $A_1av(r_j)$ is LOW) and (the density of the neighboring area of neurons is HIGH or MODERATE) (i.e. there are other prototypical nodes that overlap with j in the input-output space; this condition apply only for some strategies of inserting rule nodes as explained below) THEN the probability of pruning node $(r_j)$ is HIGH. The above pruning rule is fuzzy and it requires that the fuzzy concepts as OLD, HIGH, etc. are predefined.

14) Aggregate rule nodes, if necessary, into a smaller number of nodes. A C-means clustering algorithm can be used for this purpose.

15) End of the *while* loop and the algorithm

The rules that represent the rule nodes need to be aggregated in clusters of rules. The degree of aggregation can vary depending on the level of granularity needed. At any time (phase) of the evolving (learning) process, fuzzy, or exact rules can be inserted and extracted [15]. Insertion of fuzzy rules is achieved through setting a new rule node for each new rule, such as the connection weights $W_1$ and $W_2$ of the rule node represent the fuzzy or the exact rule. The process of rule extraction can be performed as aggregation of several rule nodes into larger hyper-spheres. For the aggregation of two-rule nodes $r_1$ and $r_2$, the following aggregation rule is used:

If $(D(W_1(r_1), W_1(r_2)) <= Thr_1)$ and $(D(W_2(r_1), W_2(r_2)) <= Thr_2)$

then aggregate $r_1$ and $r_2$ into $r_{agg}$ and calculate the centres of the new rule node as:

$W_1(r_{agg}) = average\ (W_1(r_1), W_1(r_2))$, $W_2(r_{agg}) = average\ (W_2(r_1), W_2(r_2))$

Here the geometrical center between two points in a fuzzy problem space is calculated with the use of an average vector operation over the two fuzzy vectors. This is based on a presumed piece-wise linear function between two points from the defined through the parameters *Sthr* and *Errthr* input and output fuzzy hyper-spheres.

## 2.3 Autoregressive Integrated Moving Average (ARIMA) Model

Given data on half hourly temperatures forecasts, a number of qualitative predictor variables, and half-hourly sequentially recorded response values of electricity demand, it seems a suitable statistical prediction model would be a time series multiple regression model. However, there was no apparent trend over time and the response variable exhibited strong autocorrelation with lag one. Besides, in Victoria, the extreme weather conditions in summer and winter months are very unpredictable. These changes occur with a very short notice, lasts a short duration, and show little regularity from year to year. These considerations and the fact that a time efficient model was desired led to the exploration of other time series statistical models.

Application of general class of statistical forecasting methods involves two basic tasks: analysis of the data series and selection of the forecasting model. A simple regression model can be represented as

$$Y = b_0 + b_1 X_1 + b_2 X_2 + \ldots\ldots + b_p X_p + e \qquad (8)$$

where $Y$ is the forecast variable, $X_1$ through $X_p$ are the explanatory variables, $b_0$ through $b_p$ are the linear regression coefficients, and $e$ the error term. If we can represent the explanatory variables as $X_1 = Y_{t-1}$, $X_2 = Y_{t-2}, \ldots X_p = Y_{t-p}$; (3) then becomes

$$Y_t = b_0 + b_1 Y_{t-1} + b_2 Y_{t-2} + \ldots\ldots + b_p Y_{t-p} + e_t \qquad (9)$$

The name Auto Regression (AR) is used to describe the above equation due to the time-lagged values of the explanatory variable. Just as it is possible to regress against past values of the series, there is a time series model, which uses past errors as explanatory variables:

$$Y_t = b_0 + b_1 e_{t-1} + b_2 e_{t-2} + \ldots\ldots + b_p e_{t-p} + e_t \qquad (10)$$

Here, explicitly, a dependence relationship is set up among the successive error terms, and the equation is called a Moving Average (MA) model. Autoregressive (AR) models can be effectively coupled with moving average (MA) models to form a general and useful class of time series models called autoregressive moving average (ARMA) models especially for stationary data. This class of models can be extended to non-stationary series by allowing differencing of the data series. These are called autoregressive integrated moving average (ARIMA) models [12]. ARIMA model building can be summarized in four steps:

- *Model Identification* Using graphs, statistics, autocorrelation function, partial autocorrelation functions, transformations, etc., achieve stationary and tentatively identify patterns and model components.
- *Parameter Estimation* Determine the model coefficients through the method of least squares, maximum likelihood methods and other techniques.
- *Model Diagnostics* Determine if the model is valid. If valid, then use the model; otherwise repeat identification, estimation, and diagnostic steps.
- *Forecast verification and Reasonableness* Having estimated an ARIMA model, it is necessary to revisit the question of identification to see if the selected model can be improved. Several statistical techniques and confidence intervals determine the validity of forecasts and track model performance top detect out of control situations.

There are such a bewildering variety of ARIMA models; it can be difficult to decide which model is most appropriate for a given set of data. Having made tentative model identification, the AR and MA parameters, seasonal and non-seasonal, have to be determined in the best possible manner. Even if the selected model appear to be best among those models considered, it is also necessary to do diagnostic checking to check that the model is adequate.

## 3. Experimentation Set-up – Training and Performance Evaluation

The data for our study were the recorded half-hourly actual electricity demand for the ten months period from January to November 1995 in the State of Victoria. Figure 3 shows a typical weekly cycle of electricity demand. Fluctuations in daily demand are prevalent with peaks occurring around midday. Extreme weather conditions in winter and summer months accentuate peaks in electricity demand due to the widespread use of electricity for heating and cooling. Other times, electricity demand is dominated primarily by ambient temperature, time of day, working or non-working day and the day of week.

The experimental system consists of two stages: modelling the prediction systems (training in the case of soft computing models) and performance evaluation. For network training, the six selected input descriptor variables were: the *minimum* and *maximum recorded temperatures*, *previous day's demand*, a value expressing the *half-hour period of the day*, *season*, and the *day of week*. To evaluate the learning capability of the soft computing models, the network was trained only on 20% of the randomly selected data. We created 3 different samples of training data to study the effect of random sampling and periodicity. Each training sample consisted of 2937 data sets representing 20% random data.

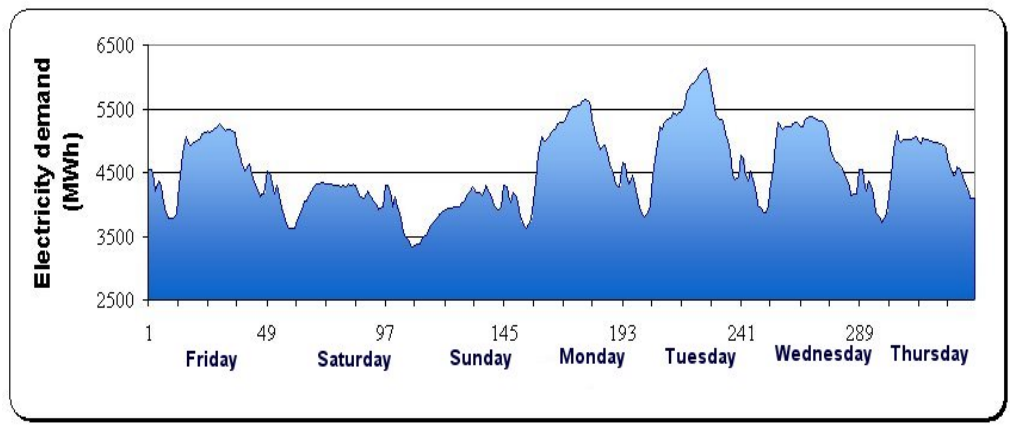

**Figure 3.** Typical weekly demand variations

Our objective is to develop an efficient forecasting model capable of producing a short-term forecast of demand for electricity. The required time-resolution of the forecast is half-hourly, and the required time-span of the forecast is 2 days. This means that the system should be able to produce a forecast of electricity demand for the next 96 time periods. The training was replicated three times using three different samples of training data and different combinations of network parameters. We used a Pentium II, 450 MHz platform for simulating the prediction models using MATLAB.

### 3.1 EFuNN training

We used 4 Gaussian membership functions for each input variable and the following evolving parameters: sensitivity threshold $Sthr$=0.99, error threshold $Errthr$=0.001 and learning rates for first and second layer = 0.05. EFuNN uses a one pass training approach. The network parameters were determined using a trial and error approach. The training was repeated three times after reinitialising the network and the worst errors were reported. Online learning in EFuNN resulted in creating 2122 rule nodes as depicted in Figure 4(b). Figure 4 (a) illustrates the EFuNN training results and the training performance is summarized in Table 1.

### 3.2 ANN training

While EFuNN is capable of adapting the architecture according to the problem, we had to perform some initial experiments to decide the architecture, activation functions and learning parameters of the artificial neural network. Our preliminary experiments helped us to formulate a feedforward neural network with 1 input layer, 2 hidden layers and an output layer [6-40-40-1]. Input layer consists of 6 neurons corresponding to the input variables. The first and second hidden layers consist of 40 neurons respectively using tanh-sigmoidal activation functions. To illustrate the convergence feature of SCGA we also trained a neural network (with same architecture) using backpropagation algorithm. To evaluate ANN learning performance, training was terminated after 2500 epochs. Training errors (RMSE) achieved after 2500 epochs for SCGA and BP are 0.0304 and 0.116 respectively. Figure 4 (c) shows the convergence of SCGA with respect to BP algorithm.

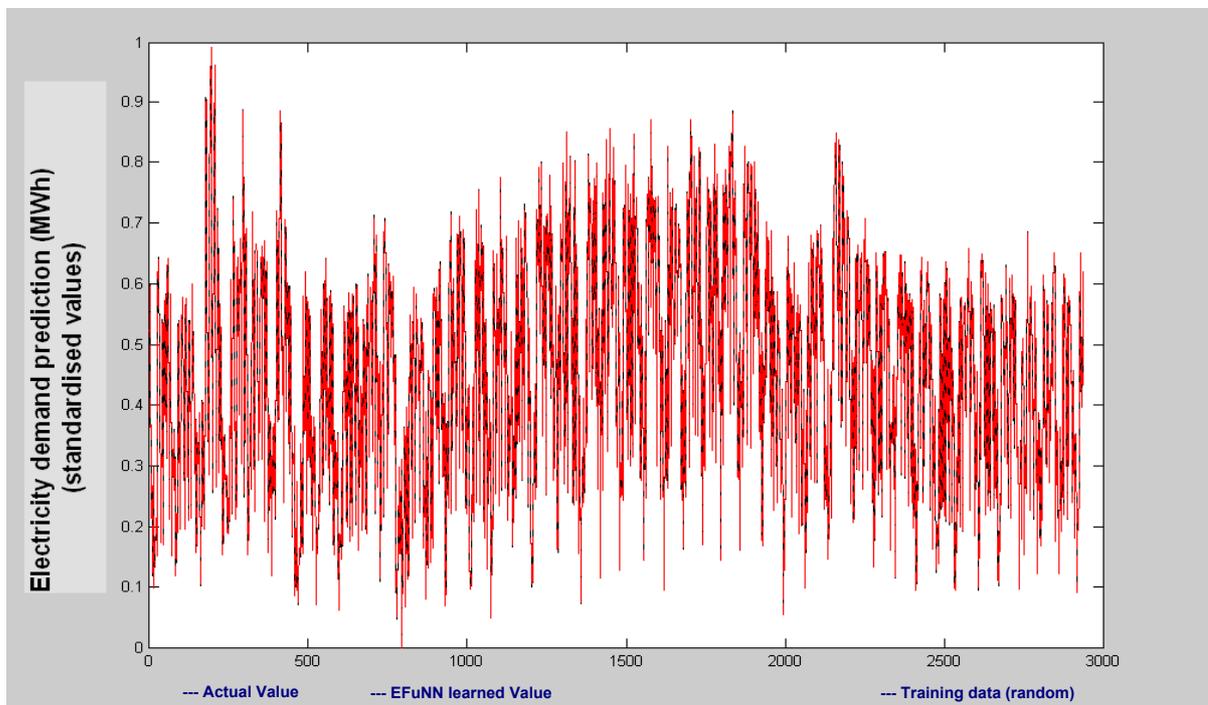

**Figure 4 (a).** Training results for EFuNN

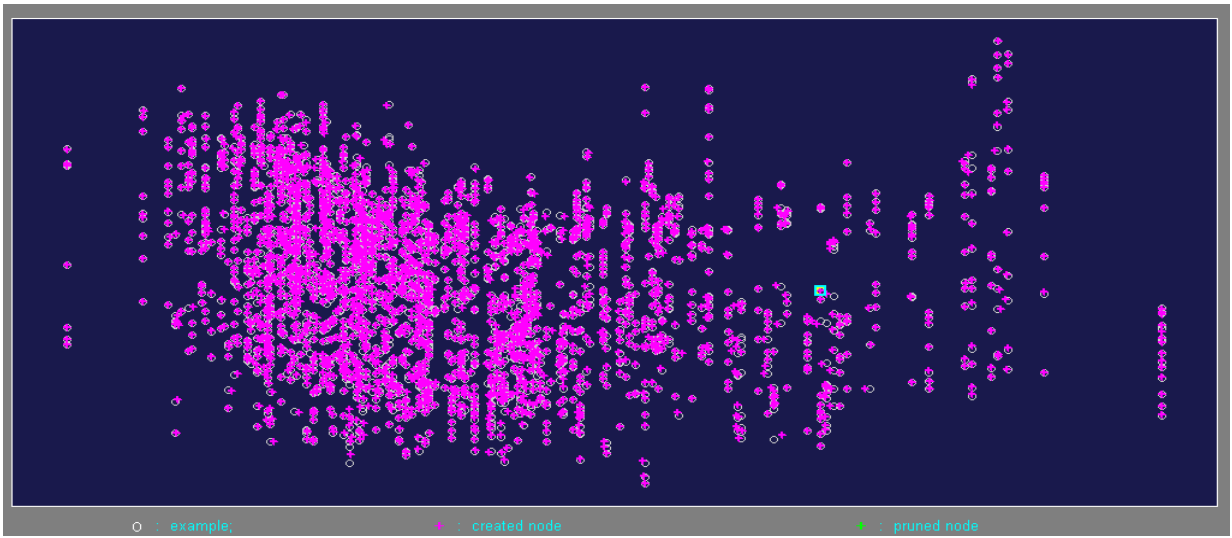

**Figure 4(b).** Learned rule nodes during EFuNN training.

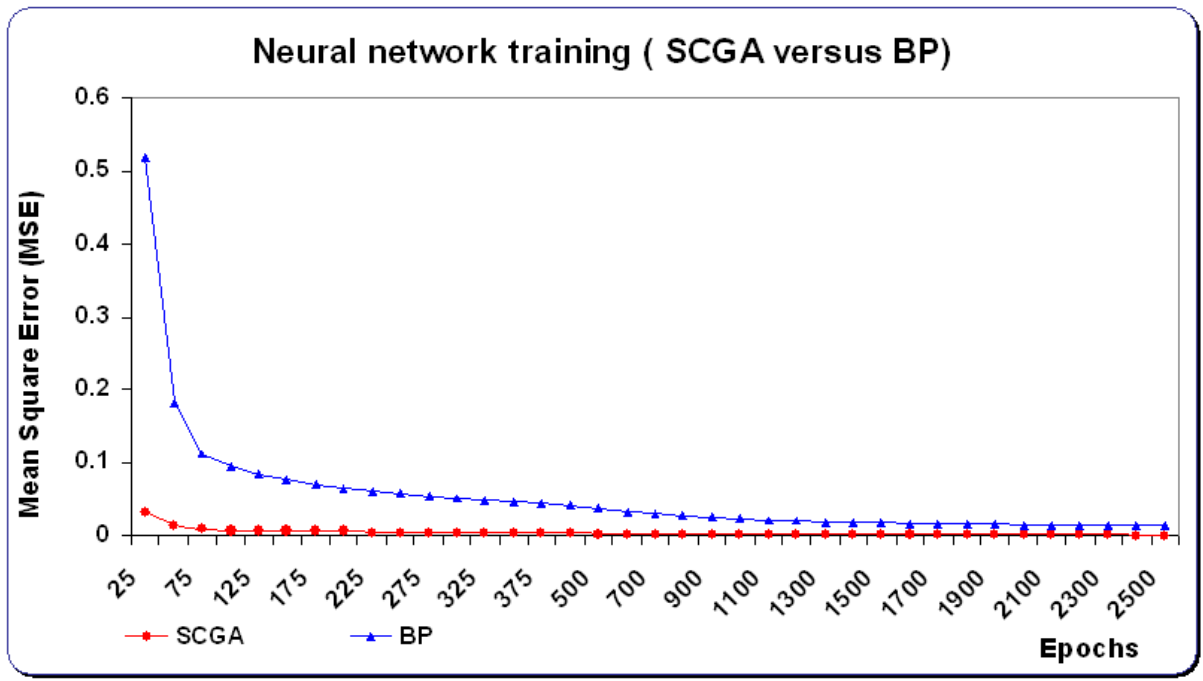

**Figure 4(c)** Comparison of ANN training using SCGA and BP

### 3.3 ARIMA model

It is observed that the electricity demand is greatly influenced by the day of the week, and the time of the day. In addition, each half-hour demand seems to depend on the previous period's demand. These three important features of the data are modelled applying Box-Jenkins ARIMA(1,1,1)(1,0,1)[48] model to data series obtained by differencing actual demand for a lag of 336 time periods.

### 3.4 Test results

Table 1 summarizes the comparative performance of evolving fuzzy neural network, artificial neural networks and ARIMA model. Figure 5 depicts the test results for prediction models considered. To have a performance evaluation the actual energy demand and the forecasts used by VHP are also plotted in

Figure 5. For more clarity, an enlarged version of Figure 5 is illustrated at the end of the paper (Figure 6).

**Table 1.** Test results and performance comparison of demand forecasting

|  | EFuNN | ANN (BP) | ANN (SCGA) | ARIMA |
|---|---|---|---|---|
| Learning epochs | 1 | 2500 | 2500 | - |
| Training error (RMSE) | 0.0013 | 0.116 | 0.0304 | - |
| Testing error (RMSE) | 0.0092 | 0.118 | 0.0323 | 0.0423 |
| Computational load (in billion flops) | 0.536 | 87.2 | 175.0 | - |

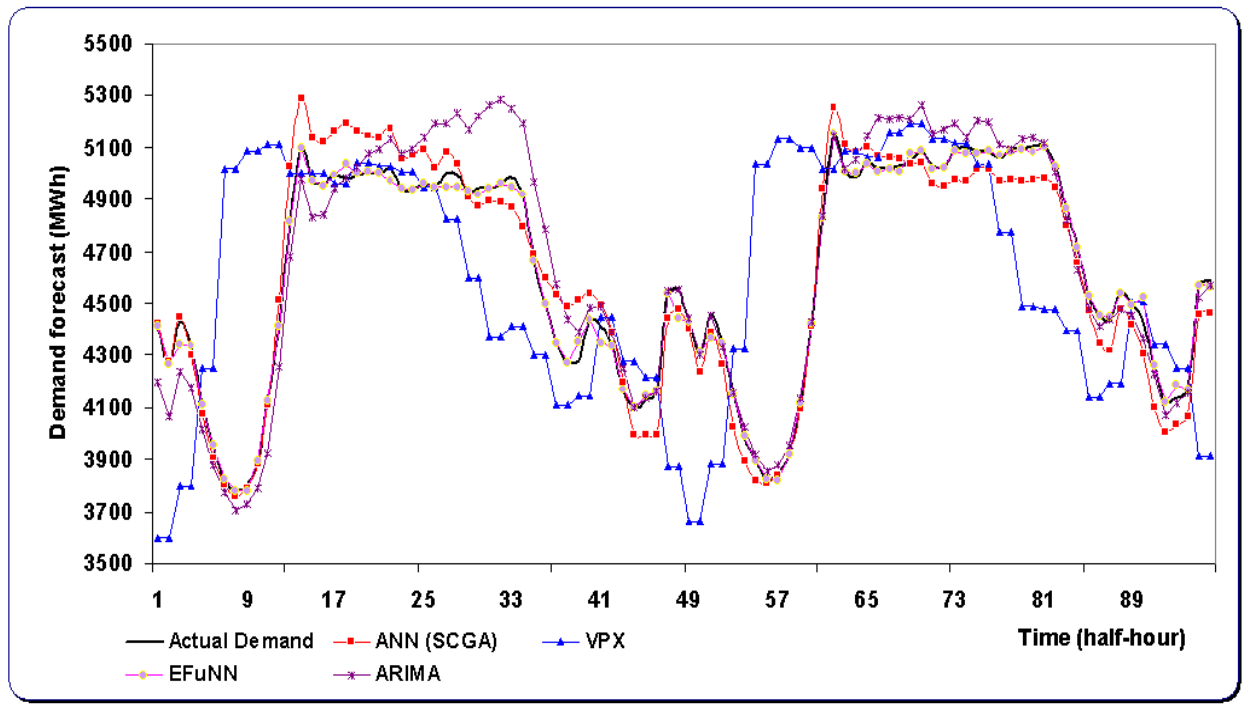

**Figure 5.** Test results and performance comparison of demand forecasts (2 days)

## 4. Discussion, Conclusions and Future Research

In this paper we attempted to model and forecast electricity demand based on two popular soft computing techniques and ARIMA model. Experimentation results reveal that EFuNN performed better than other techniques in terms of low RMSE error and less computational load (less performance time). All the considered methods performed significantly better than the forecasts used by VPX. Conjugate gradient algorithms normally converge faster than the steepest descent. As depicted in Figure 4 (C) our experimentation results reveal that ANN trained by SCGA converged much faster than BP algorithm. Alternatively, BP training needs more epochs (longer training time) to achieve better performance. The soft computing models considered on the other hand are easy to implement and produces desirable mapping function by training on the given data set. All the soft computing models require information only about the input variables for generating forecasts thereby reducing the tedious analysis and trail and error methods as in the case of ARIMA model.

Evolving fuzzy neural network makes use of the linguistic knowledge of fuzzy inference system and the learning capability of neural networks. Hence the neuro-fuzzy system is able to precisely model the uncertainty and imprecision within the data as well as to incorporate the learning ability of neural networks. Even though the performance of neuro-fuzzy systems is dependent on the problem domain, very often the results are better while compared to pure neural network approach [3] [11]. Compared to neural networks, an important advantage of neuro-fuzzy systems is its reasoning ability (*if-then* rules) of any particular state. A fully trained EFuNN could be replaced by a set of *if-then* rules [15]. A simple example of a learned EFuNN learned rule is illustrated below.

" If the *maximum temperature* of the day is HIGH **and** *minimum temperature* of the day is LOW **and** *previous days demand* is MEDIUM **and** it is *summer* (HIGH) **and** *9.00 AM* (HIGH) **and** a *Monday* (HIGH) **then** the *electricity demand* is MEDIUM."

EFuNN uses a hybrid learning technique (a mixture of unsupervised and supervised learning) to fine-tune the parameters of the fuzzy inference system. As EFuNN adopts a single pass training (1 epoch) it is more adaptable and easy for further online training which might be highly useful for online forecasting and bidding. Another important feature of EFuNN is that the user has the flexibility to construct the network (by selecting the parameters). Hence for applications where speed is more important than the accuracy a faster network can be selected. However an important disadvantage of EFuNN is the determination of the network parameters like number and type of membership functions for each input variable, sensitivity threshold, error threshold and the learning rates. Even though a trial and error approach is practical, when the problem becomes complicated (large number of input variables) determining the optimal parameters will be a tedious task.

It is interesting to note the RMSE of ARIMA model on the test set which is approximately 180% lesser than the neural network trained using backpropagation algorithm (2500 epochs training). The ANN performance (using BP) could have been improved if we had increased the number of epochs. Our experiments on three separate data samples reveal that the results are not dependent on the data sample. We used only 20% of the total data to evaluate the learning capability of the soft computing models. Network performance could have been further improved by providing more training data. Another interesting fact about the considered soft computing models are their robustness and capability to handle noisy and approximate data that are typical in power systems, and therefore should be more reliable in worst situations.

The important drawback with the conventional design of ANN is that the designer has to specify the number of neurons, their distribution over several layers, interconnection between them, initial weights, type of learning algorithm and parameters. In this paper, we have considered a forecasting time resolution of 30 minutes. Considering the rapid fluctuation of demand and for a more effective energy management, one might have to consider even a low resolution forecasting (e.g. 5 minutes). Our future works include using the Adaptive Learning by Evolutionary Computation (ALEC) framework [14] for optimising the neural networks used to build forecasting models. Similar procedures might be used to automatically adapt the optimal combination of network parameters for the evolving fuzzy neural network.

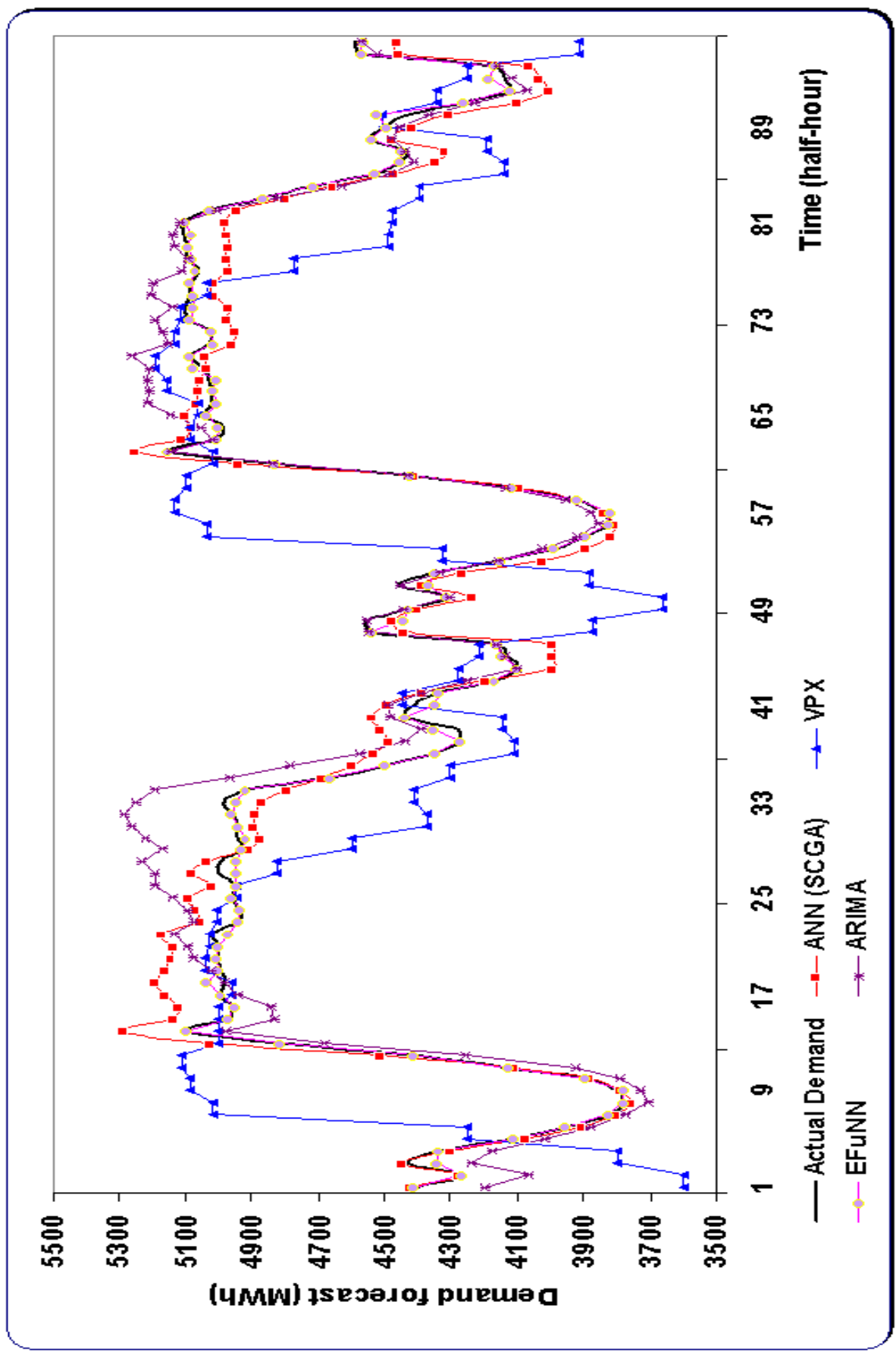

**Figure 6.** Test results and performance comparison of demand forecasts (2 days)